\pdfoutput=1
\documentclass[11pt,a4paper]{article}
\usepackage[hyperref]{naaclhlt2018}

\usepackage{times}
\usepackage{url}
\usepackage{latexsym}

\usepackage[british]{babel}
\usepackage{graphicx}
\usepackage{colortbl}
\usepackage{hhline}
\usepackage{url}
\usepackage{latexsym}
\usepackage{amssymb}
\usepackage{amsmath}
\usepackage{enumerate}
\usepackage{algorithm}
\usepackage{algorithmicx}
\usepackage[noend]{algpseudocode}
\usepackage{booktabs}
\usepackage{proof}
\usepackage{paralist}
\usepackage{tikz-dependency}
\usepackage{tikz} 
\usepackage{graphicx, subfigure}
\usepackage{epstopdf}

\usepackage{caption}
\usepackage[latin1]{inputenc}

\newcommand{\eat}[1]{}

\makeatletter
\useshorthands{"}%
\defineshorthand{"-}{\nobreak-\bbl@allowhyphens}
\makeatother

\newcommand{\transname}[1]{\ensuremath{\mathsf{#1}}}

\newcommand{\stacktop}{{\mid}}

\mathchardef\mhyphen="2D 

\newcommand{\la}{\transname{Left\mhyphen Arc}}
\newcommand{\ra}{\transname{Right\mhyphen Arc}}
\newcommand{\na}{\transname{No\mhyphen Arc}}
\newcommand{\sh}{\transname{Shift}}

\aclfinalcopy 
\def\aclpaperid{743} 

\title{Non-Projective Dependency Parsing with Non-Local Transitions}

\author{Daniel Fern\'{a}ndez-Gonz\'{a}lez \and Carlos G\'{o}mez-Rodr\'{i}guez\\
	Universidade da Coru\~{n}a\\
	FASTPARSE Lab, LyS Research Group, Departamento de Computaci\'{o}n \\
	Campus de Elvi\~{n}a, s/n, 15071 A Coru\~{n}a, Spain \\
  {\tt d.fgonzalez@udc.es}, {\tt carlos.gomez@udc.es}\\}

\date{}

\begin{document}
\maketitle
\begin{abstract}
We present a novel transition system, based on the Covington non-projective parser, introducing non-local transitions that can directly create arcs involving nodes to the left of the current focus positions. This avoids the need for long sequences of $\na$ transitions to create long-distance arcs, thus alleviating error propagation. The resulting parser outperforms the original version and achieves the best accuracy on the Stanford Dependencies conversion of the Penn Treebank among greedy transition-based parsers.

\end{abstract}

\section{Introduction}

Greedy transition-based parsers are popular in NLP, as they provide competitive accuracy with high efficiency. They syntactically analyze a sentence by greedily applying transitions, which read it from left to right and produce a dependency tree.

However, this greedy process is prone to error propagation: one wrong choice of transition can lead the parser to an erroneous state, causing more incorrect decisions. This is especially crucial for long attachments requiring a larger number of transitions. In addition, transition-based parsers traditionally focus on only two words of the sentence and their local context to choose the next transition. The lack of a global perspective favors the presence of errors when creating arcs involving multiple transitions. As expected, transition-based parsers build short arcs more accurately than long ones \cite{mcdonald07emnlp}.

Previous research such as \cite{buffertrans} and \cite{Qi2017} proves that the widely-used projective \textit{arc-eager} transition-based parser of \newcite{Nivre2003} benefits from shortening the length of transition sequences by creating non-local attachments. In particular, they augmented the original transition system with new actions whose behavior entails more than one arc-eager transition and involves a context beyond the traditional two focus words. \newcite{attardi06} and \newcite{sartorio13} also extended the \textit{arc-standard} transition-based algorithm \cite{nivre04acl} with the same success.

In the same vein, we present a novel unrestricted non-projective transition system based on the well-known algorithm by \newcite{covington01fundamental} that shortens the transition sequence necessary to parse a given sentence by the original algorithm, which becomes linear instead of quadratic with respect to sentence length. To achieve that, we propose new transitions that affect non-local words and are equivalent to one or more Covington actions, in a similar way to the transitions defined by \newcite{Qi2017} based on the arc-eager parser. Experiments show that this novel variant significantly outperforms the original one in all datasets tested, and
achieves the best reported accuracy for a greedy dependency parser on the Stanford Dependencies conversion of the WSJ Penn Treebank.
 
\section{Non-Projective Covington Parser}
\label{sec:cov}

The original non-projective parser defined by \newcite{covington01fundamental} was modelled under the transition-based parsing framework by \newcite{Nivre2008}. We only sketch this transition system briefly for space reasons, and refer to \cite{Nivre2008} for details.

\begin{figure*}
\small
\begin{tabbing}
\hspace{2cm}\=\hspace{2.4cm}\= \kill
\emph{Covington:} \> \sh: \> $\langle {\lambda_1}, {\lambda_2}, {j \stacktop B} , {A} \rangle
\Rightarrow \langle {\lambda_1} \cdot {\lambda_2 \stacktop j}, [], B , A \rangle$ { }\\[1mm]
\> \na: \> $\langle {\lambda_1 \stacktop i}, {\lambda_2}, B , A \rangle
\Rightarrow \langle \lambda_1, {i \stacktop \lambda_2  }, B , A \rangle$ { } \\[1mm]
\> \la: \> $\langle {\lambda_1 \stacktop i} , {\lambda_2}, {j \stacktop B} ,
A \rangle \Rightarrow
\langle \lambda_1, {i \stacktop \lambda_2}, {j \stacktop B} , A \cup \{ j \rightarrow i \} \rangle$\\
\> \> \small{only if $\nexists x \mid x \rightarrow i \in A$ (single-head) and $i  \rightarrow^\ast j \not\in A$ (acyclicity).}\\[1mm]
\> \ra: \> $\langle {\lambda_1 \stacktop i} , {\lambda_2}, {j \stacktop B} ,
A \rangle \Rightarrow
\langle \lambda_1, {i \stacktop \lambda_2}, {j \stacktop B} , A \cup \{ i \rightarrow j \} \rangle$\\
\> \> \small{only if $\nexists x \mid x \rightarrow j \in A$ (single-head) and $j  \rightarrow^\ast i \not\in A$ (acyclicity).}{}
\end{tabbing}

\vspace{1mm}
\begin{tabbing}
\hspace{2cm}\=\hspace{2.4cm}\= \kill
\emph{NL-Covington:} \> \sh: \> $\langle {\lambda_1}, {\lambda_2}, {j \stacktop B} , {A} \rangle
\Rightarrow \langle {\lambda_1} \cdot {\lambda_2 \stacktop j}, [], B , A \rangle$ { }\\[1mm]
\> \la$_k$: \> $\langle {\lambda_1 \stacktop i_k \stacktop ... \stacktop i_1} , {\lambda_2}, {j \stacktop B} ,
A \rangle \Rightarrow
\langle \lambda_1, {i_k \stacktop ... \stacktop i_1 \stacktop \lambda_2}, {j \stacktop B} , A \cup \{ j \rightarrow i_k \} \rangle$\\
\> \> \small{only if $\nexists x \mid x \rightarrow i_k \in A$ (single-head) and $i_k  \rightarrow^\ast j \not\in A$ (acyclicity).}\\[1mm]
\> \ra$_k$: \> $\langle {\lambda_1 \stacktop i_k \stacktop ... \stacktop i_1} , {\lambda_2}, {j \stacktop B} ,
A \rangle \Rightarrow
\langle \lambda_1, {i_k \stacktop ... \stacktop i_1 \stacktop \lambda_2}, {j \stacktop B} , A \cup \{ i_k \rightarrow j \} \rangle$\\
\> \> \small{only if $\nexists x \mid x \rightarrow j \in A$ (single-head) and $j  \rightarrow^\ast i_k \not\in A$ (acyclicity).}{}
\end{tabbing}

\caption{Transitions of the non-projective Covington (top) and NL-Covington (bottom) dependency parsers. The notation $i \rightarrow^\ast j \in A$ means that there is a (possibly empty) directed path from $i$ to $j$ in $A$.}
\label{fig:transitionsfusion}
\end{figure*}

Parser configurations have the form {$c=\langle {\lambda_1} , {\lambda_2} , {B} , {A} \rangle$}, where $\lambda_1$ and $\lambda_2$ are lists of partially processed words, $B$ a list (called buffer) of unprocessed words, and $A$ the set of dependency arcs built so far. Given an input string $w_1 \cdots w_n$, the parser starts at the initial configuration $c_s(w_1 \ldots w_n) = \langle [],[],[1 \ldots n],\emptyset \rangle$ and runs transitions until a terminal configuration of the form $ \langle \lambda_1, \lambda_2 , [] , A \rangle$ is reached: at that point, $A$ contains the dependency graph for the input.\footnote{Note that, in general, $A$ is a forest, but it can be converted to a tree by linking headless nodes as dependents of an artificial root node at position $0$.}

\begin{figure}[t]
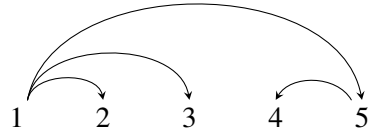

 \begin{center}
 \begin{dependency}[theme = simple]
 \begin{deptext}[column sep=2em]
 1 \& 2 \& 3 \& 4 \& 5 \\
 \end{deptext}
 \depedge{1}{2}{}
 \depedge{1}{5}{}
 \depedge[arc angle=60]{5}{4}{}
 \depedge[arc angle=80]{1}{3}{}
 \end{dependency}
 \end{center}
 \caption{Dependency tree for an input sentence.}
 \label{fig:exdeptree}       
 \end{figure}

The set of transitions is shown in the top half of Figure \ref{fig:transitionsfusion}. Their logic can be summarized as follows: when in a configuration of the form $\langle {\lambda_1 \stacktop i} , {\lambda_2}, {j \stacktop B} , A \rangle$, the parser has the chance to create a dependency involving words $i$ and $j$, which we will call left and right focus words of that configuration. The $\la$ and $\ra$ transitions are used to create a leftward ($i \leftarrow j$) or rightward arc ($i \rightarrow j$), respectively, between these words, and also move $i$ from $\lambda_1$ to the first position of $\lambda_2$, effectively moving the focus to $i-1$ and $j$. If no dependency is desired between the focus words, the $\na$ transition makes the same modification of $\lambda_1$ and $\lambda_2$, but without building any arc. Finally, the $\sh$ transition moves the whole content of the list $\lambda_2$ plus $j$ to $\lambda_1$ when no more attachments are pending between $j$ and the words of $\lambda_1$, thus reading a new input word and placing the focus on $j$ and $j+1$. Transitions that create arcs are disallowed in configurations where this would violate the single-head or acyclicity constraints (cycles and nodes with multiple heads are not allowed in the dependency graph). Figure \ref{fig:seq} shows the transition sequence in the Covington transition system which derives the dependency graph in Figure \ref{fig:exdeptree}.

\begin{figure}
 \begin{footnotesize}
 \begin{center}
 \vspace*{13pt}
 \begin{tabular}{@{\hskip 0pt}l@{\hskip 3pt}c@{\hskip 8pt}c@{\hskip 8pt}c@{\hskip 8pt}c@{\hskip 3pt}}
 \hline\noalign{\smallskip}
 Tran. & $\lambda_1$ & $\lambda_2$ & Buffer & Arc \\
 \noalign{\smallskip}\hline\noalign{\smallskip}
  & [ ] & [ ] & [ 1, 2, 3, 4, 5 ] &  \\
 \textsc{SH} & [ 1 ] & [ ] & [ 2, 3, 4, 5 ] &  \\
 \textsc{RA} & [ ] & [ 1 ] & [ 2, 3, 4, 5 ] & \scriptsize{$1 \rightarrow 2$} \\
 \textsc{SH} & [ 1, 2 ] & [ ] & [ 3, 4, 5 ] &  \\
 \textsc{NA} & [ 1 ] & [ 2 ] & [ 3, 4, 5 ] &  \\
 \textsc{RA} & [ ] & [ 1, 2 ] & [ 3, 4, 5 ] & \scriptsize{$1 \rightarrow 3$} \\
 \textsc{SH} & [ 1, 2, 3 ] & [ ] & [ 4, 5 ] &  \\
 \textsc{SH} & [ 1, 2, 3, 4 ] & [ ] & [ 5 ] &  \\
 \textsc{LA} & [ 1, 2, 3 ] & [ 4 ] & [ 5 ] &   \scriptsize{$4 \leftarrow 5$} \\
 \textsc{NA} & [ 1, 2] & [ 3, 4 ] & [ 5 ] &  \\
 \textsc{NA} & [ 1 ] & [ 2, 3, 4 ] & [ 5 ] &  \\
 \textsc{RA} & [ ] & [ 1, 2, 3, 4 ] & [ 5 ] & \scriptsize{$1 \rightarrow 5$} \\
 \textsc{SH} & [ 1, 2, 3, 4, 5 ] & [ ] & [ ] & \\
 \noalign{\smallskip}\hline
 \end{tabular}
 \caption{Transition sequence for parsing the sentence in Figure~\ref{fig:exdeptree} using
 the Covington parser (LA=\textsc{Left-Arc}, RA=\textsc{Right-Arc}, NA=\textsc{No-Arc}, SH=\textsc{Shift}).} \label{fig:seq}       
 \vspace*{13pt}
 \end{center}
 \end{footnotesize}
\end{figure}

The resulting parser can generate arbitrary non-projective trees, and its complexity is $O(n^2)$.
\section{Non-Projective NL-Covington Parser}

The original logic described by \newcite{covington01fundamental} parses a sentence by systematically traversing every pair of words. The $\sh$ transition, introduced by \newcite{Nivre2008} in the transition-based version, is an optimization that avoids the need to apply a sequence of $\na$ transitions to empty the list $\lambda_1$ before reading a new input word.

However, there are still situations where sequences of $\na$ transitions are needed. For example, if we are in a configuration $C$ with focus words $i$ and $j$ and the next arc we need to create goes from $j$ to $i-k$ $(k>1)$, then we will need $k-1$ consecutive $\na$ transitions to move the left focus word to $i$ and then apply $\la$. This could be avoided if a non-local $\la$ transition could be undertaken directly at $C$, creating the required arc and moving $k$ words to $\lambda_2$ at once. The advantage of such approach would be twofold: (1) less risk of making a mistake at $C$ due to considering a limited local context, and (2) shorter transition sequence, alleviating error propagation.

We present a novel transition system called \textit{NL-Covington} (for ``non-local Covington''), described in the bottom half of Figure~\ref{fig:transitionsfusion}. It consists in a modification of the non-projective Covington algorithm where:
(1) the $\la$ and $\ra$ transitions are parameterized with $k$, allowing the immediate creation of any attachment between $j$ and the $k$th leftmost word in $\lambda_1$ and moving $k$ words to $\lambda_2$ at once, and (2) the $\na$ transition is removed since it is no longer necessary.

This new transition system can use some restricted global information to build non-local dependencies and, consequently, reduce the number of transitions needed to parse the input. For instance, as presented in Figure~\ref{fig:seq2}, the NL-Covington parser will need 9 transitions, 
instead of 12 traditional Covington actions, 
to analyze the sentence in Figure~\ref{fig:exdeptree}. 

In fact, while in the standard Covington algorithm a transition sequence for a sentence of length $n$ has length $O(n^2)$ in the worst case (if all nodes are connected to the first node, then we need to traverse every node to the left of each right focus word); for NL-Covington the sequence length is always $O(n)$: one $\sh$ transition for each of the $n$ words, plus one arc-building transition for each of the $n-1$ arcs in the dependency tree.
Note, however, that this does not affect the parser's time complexity, which is still quadratic as in the original Covington parser. This is because the algorithm has $O(n)$ possible transitions to be scored at each configuration, while the original Covington has $O(1)$ transitions due to being limited to creating local leftward/rightward arcs between the focus words.

The completeness and soundness of NL-Covington can easily be proved as there is a mapping between transition sequences of both parsers, where a sequence of $k-1$ $\na$ and one arc transition in Covington is equivalent to a $\la_k$ or $\ra_k$ in NL-Covington.

 \begin{figure}
 \begin{center}
 \footnotesize
 \begin{tabular}{@{\hskip 0pt}l@{\hskip 3pt}c@{\hskip 8pt}c@{\hskip 8pt}c@{\hskip 8pt}c@{\hskip 3pt}}
 \hline\noalign{\smallskip}
 Tran. & $\lambda_1$ & $\lambda_2$ & Buffer & Arc \\
 \noalign{\smallskip}\hline\noalign{\smallskip}
  & [ ] & [ ] & [ 1, 2, 3, 4, 5 ] &  \\
 \textsc{SH} & [ 1 ] & [ ] & [ 2, 3, 4 , 5 ] &  \\
 \textsc{RA$_1$} & [ ] & [ 1 ] & [ 2, 3, 4 , 5 ] & \scriptsize{$1 \rightarrow 2$} \\
 \textsc{SH} & [ 1, 2 ] & [ ] & [ 3, 4, 5 ] &  \\
 \textsc{RA$_2$} & [ ] & [ 1, 2 ] & [ 3, 4, 5 ] & \scriptsize{$1 \rightarrow 3$} \\
 \textsc{SH} & [ 1, 2, 3 ] & [ ] & [ 4, 5 ] &  \\
 \textsc{SH} & [ 1, 2, 3, 4 ] & [ ] & [ 5 ] &  \\
 \textsc{LA$_1$} & [ 1, 2, 3 ] & [ 4 ] & [ 5 ] &   \scriptsize{$4 \leftarrow 5$} \\
 \textsc{RA$_3$} & [ ] & [ 1, 2, 3, 4 ] & [ 5 ] & \scriptsize{$1 \rightarrow 5$} \\
 \textsc{SH} & [ 1, 2, 3, 4, 5 ] & [ ] & [ ] & \\
 \noalign{\smallskip}\hline
 \end{tabular}
 \caption{Transition sequence for parsing the sentence in Figure~\ref{fig:exdeptree} using
 the NL-Covington parser (LA=\textsc{Left-Arc}, RA=\textsc{Right-Arc}, SH=\textsc{Shift}).} \label{fig:seq2}       
 \vspace*{13pt}
 \end{center}
 \end{figure}

\section{Experiments}
\subsection{Data and Evaluation}
We use 9 datasets\footnote{We excluded the languages from CoNLL-X that also appeared in CoNLL-XI, i.e., if a language was present in both shared tasks, we used the latest version.} from the CoNLL-X \cite{buchholz06} and all datasets from the CoNLL-XI shared task \cite{conll2007}. To compare our system to the current state-of-the-art transition-based parsers, we also evaluate it on the Stanford Dependencies \cite{deMarneffe2008} conversion (using the Stanford parser v3.3.0)\footnote{\url{https://nlp.stanford.edu/software/lex-parser.shtml}} of the WSJ Penn Treebank \cite{marcus93}, hereinafter PT-SD, with standard splits. Labelled and Unlabelled Attachment Scores (LAS and UAS) are computed excluding punctuation only on the PT-SD, for comparability. 
We repeat each experiment with three independent random initializations and report the average accuracy. Statistical significance is assessed by a paired test with 10,000 bootstrap samples.

\subsection{Model}
To implement our approach we take advantage of the model architecture described in \newcite{Qi2017} for the \textit{arc-swift} parser, which extends the architecture of \newcite{Kiperwasser2016} by applying a biaffine combination during the featurization process. We implement both the Covington and NL-Covington parsers under this architecture, adapt the featurization process with biaffine combination of \newcite{Qi2017} to these parsers, and use their same training setup. More details about these model parameters are provided in Appendix \ref{app:modeldescription}. 

Since this architecture uses batch training, we train with a static oracle. 
The NL-Covington algorithm has no spurious ambiguity at all, so there is only one possible static oracle: canonical transition sequences are generated by choosing the transition that builds the shortest pending gold arc involving the current right focus word $j$, or $\sh$ if there are no unbuilt gold arcs involving $j$.

We note that a dynamic oracle can be obtained 
for the NL-Covington parser by adapting the one for standard Covington of \newcite{GomFerACL2015}. As NL-Covington transitions are concatenations of Covington ones, their loss calculation algorithm is compatible with NL-Covington. Apart from error exploration, this also opens the way to incorporating non-monotonicity \cite{FerGomACL2017}. While these approaches have shown to improve accuracy under online training settings, here we prioritize homogeneous comparability to \cite{Qi2017}, so we use batch training and a static oracle, and still obtain state-of-the-art accuracy for a greedy parser.

\begin{table}
\begin{small}
\begin{center}
\centering
\begin{tabular}{@{\hskip 0pt}l|cc|cc@{\hskip 0pt}}
& \multicolumn{2}{c|}{Covington}
& \multicolumn{2}{c}{NL-Covington}
\\
Language
& UAS & LAS
& UAS & LAS 
\\
\hline
Arabic & 66.67 & 53.24 & \textbf{68.69}  & \textbf{54.59} \\ 
Basque &  74.31 & 66.18 & \textbf{75.45} & \textbf{67.61}  \\ 
Catalan & 91.93 & 86.12 & \textbf{92.60} & \textbf{86.99} \\ 
Chinese & 83.87 & 76.19 & \textbf{85.25} & \textbf{77.56}   \\  
Czech & 84.27 & 77.91 & \textbf{86.26}  & \textbf{79.95} \\ 
English & 89.94 & 88.74  & \textbf{91.51}  &  \textbf{90.47} \\  
Greek & 79.91 & 72.65  & \textbf{80.61} & \textbf{73.41}  \\ 
Hungarian & 76.80 & 65.21 & \textbf{78.57} & \textbf{67.51}  \\ 
Italian & 82.03 & 75.87 & \textbf{83.63} & \textbf{78.03}   \\ 
Turkish & 80.29 & 70.68 & \textbf{81.30} & \textbf{71.28} \\ 
\hline
Bulgarian & 81.78 & 76.23 & \textbf{83.65} & \textbf{78.40} \\ 
Danish & 86.56 & 81.18 & \textbf{88.40} & \textbf{82.77} \\ 
Dutch & 86.19 & 82.24 & \textbf{87.45} & \textbf{83.76} \\ 
German &  85.72 & 82.28 & \textbf{87.24} & \textbf{83.92}  \\ 
Japanese & 92.20 & 90.41 & \textbf{93.63} & \textbf{91.65}   \\ 
Portuguese & 86.69 & 82.19 & \textbf{87.89} & \textbf{83.69} \\ 
Slovene & 76.07 & 66.81 &  \textbf{77.83} & \textbf{69.74}   \\ 
Spanish & 74.67 & 69.41 & \textbf{76.58} & \textbf{71.60} \\ 
Swedish &  74.65 & 64.67 & \textbf{75.62} & \textbf{65.95}  \\ 
\hline
Average & 81.82  & 75.17 & \textbf{83.27} & \textbf{76.78}  \\
\hline
\multicolumn{5}{c}{}\\
\end{tabular}
\centering
\caption{Parsing accuracy (UAS and LAS, including punctuation) of the Covington  and NL-Covington non-projective parsers on CoNLL-XI (first block) and CoNLL-X (second block) datasets. Best results for each language are shown in bold. All improvements in this table are statistically significant ($\alpha = .05$).}
\label{tab:results}
\end{center}
\end{small}
\end{table}

\subsection{Results}
Table~\ref{tab:results} presents a comparison between the Covington parser and the novel variant developed here. The NL-Covington parser outperforms the original version in all datasets tested, with all improvements statistically significant ($\alpha = .05$).

\begin{table}
\begin{small}
\begin{center}
\centering
\begin{tabular}{@{\hskip 0pt}lccc@{\hskip 0pt}}
Parser & Type & UAS & LAS \\
\hline
\cite{CheMan2014} & gs &  91.8  &  89.6  \\
\cite{Dyer2015} & gs & 93.1 & 90.9  \\
\cite{Weiss2015} greedy  & gs & 93.2  &  91.2  \\
\cite{Ballesteros2016}  & gd &  93.5  &  91.4  \\
\cite{Kiperwasser2016} & gd & 93.9 & 91.9 \\
\cite{Qi2017} & gs & 94.3 & 92.2 \\
\textbf{This work} & gs &  \textbf{94.5} &  \textbf{92.4}  \\
\hline
\cite{Weiss2015} beam  & b(8) & 94.0  &  92.1  \\
\cite{Alberti2015}  & b(32) & 94.2  &  92.4  \\
\cite{Andor2016}  & b(32) & 94.6  &  92.8  \\
\cite{Shi2017}  & dp & 94.5  &  -  \\
\hline
\cite{Kuncoro2016} (constit.) & c & 95.8  &  94.6  \\
 
\hline
\multicolumn{1}{c}{}\\
\end{tabular}
\centering
\caption{Accuracy comparison of state-of-the-art transition-based dependency parsers on PT-SD. The ``Type'' column shows the type of parser: \emph{gs} is a greedy parser trained with a static oracle, \emph{gd} a greedy parser trained with a dynamic oracle, \emph{b(n)} a beam search parser with beam size $n$, \emph{dp} a parser that employs global training with dynamic programming, and \emph{c} a constituent parser with conversion to dependencies.
}
\label{tab:comparison}
\end{center}
\end{small}
\end{table}

Table~\ref{tab:comparison} compares our novel system with other state-of-the-art transition-based dependency parsers on the PT-SD. Greedy parsers are in the first block, beam-search and dynamic programming parsers in the second block. The third block shows the best result on this benchmark, obtained with constituent parsing with generative re-ranking and conversion to dependencies.
Despite being the only non-projective parser tested on a practically projective dataset,\footnote{Only 41 out of 39,832 sentences of the PT-SD training dataset present some kind of non-projectivity.} our parser achieves the highest score among greedy transition-based models (even above those trained with a dynamic oracle). 

We even slightly outperform the arc-swift system of \newcite{Qi2017}, with the same model architecture, implementation and training setup, but based on the projective arc-eager transition-based parser instead. This may be because our system takes into consideration any permissible attachment between the focus word $j$ and any word in $\lambda_1$ at each configuration, while their approach is limited by the arc-eager logic: it allows all possible rightward arcs (possibly fewer than our approach as the arc-eager stack usually contains a small number of words), but only one leftward arc is permitted per parser state. It is also worth noting that the arc-swift and NL-Covington parsers have the same worst-case time complexity, ($O(n^2)$), as adding non-local arc transitions to the arc-eager parser increases its complexity from linear to quadratic, but it does not affect the complexity of the Covington algorithm. Thus, it can be argued that this technique is better suited to Covington than to arc-eager parsing.

We also compare NL-Covington to the arc-swift parser on the CoNLL datasets (Table~\ref{tab:results2}). For fairness of comparison, we projectivize (via maltparser\footnote{\url{http://www.maltparser.org/}}) all training datasets, instead of filtering non-projective sentences, as some of the languages are significantly non-projective. Even doing that, the NL-Covington parser improves over the arc-swift system in terms of UAS in 14 out of 19 datasets, 
obtaining statistically significant improvements in accuracy on 7 of them, and statistically significant decreases in just one.

\begin{table}
\begin{small}
\begin{center}
\centering
\begin{tabular}{@{\hskip 0pt}l|cc|cc@{\hskip 0pt}}
& \multicolumn{2}{c|}{Arc-swift}
& \multicolumn{2}{c}{NL-Covington}
\\
Language
& UAS & LAS
& UAS & LAS 
\\
\hline
Arabic & 67.54 & 53.65 & \textbf{68.69}$^*$  & \textbf{54.59}$^*$ \\ 
Basque &  74.88 & 67.44 & \textbf{75.45} & \textbf{67.61}  \\ 
Catalan & \textbf{92.98} & \textbf{87.51}$^*$ & 92.60 & 86.99 \\ 
Chinese & 84.96 & 77.34 & \textbf{85.25} & \textbf{77.56}   \\  
Czech & 85.92 & 79.82 & \textbf{86.26}  & \textbf{79.95} \\ 
English & 91.41 & 90.43  & \textbf{91.51}  &  \textbf{90.47} \\  
Greek & \textbf{81.64}$^*$ & \textbf{74.56}$^*$  & 80.61 & 73.41  \\ 
Hungarian & \textbf{78.70} & \textbf{69.27}$^*$ & 78.57 & 67.51  \\ 
Italian & 83.29 & \textbf{78.60}$^*$ & \textbf{83.63} & 78.03   \\ 
Turkish & 79.56 & 70.22 & \textbf{81.30}$^*$ & \textbf{71.28}$^*$ \\ 
\hline
Bulgarian & 83.28 & 78.19 & \textbf{83.65} & \textbf{78.40} \\ 
Danish & 87.86 & 82.58 & \textbf{88.40}$^*$ & \textbf{82.77} \\ 
Dutch & 83.27 & 80.14 & \textbf{87.45}$^*$ & \textbf{83.76}$^*$ \\ 
German &  86.28 & 82.97 & \textbf{87.24}$^*$ & \textbf{83.92}$^*$  \\ 
Japanese & \textbf{93.64} & \textbf{91.92} & 93.63 & 91.65   \\ 
Portuguese & 87.01 & 83.09 & \textbf{87.89}$^*$ & \textbf{83.69}$^*$ \\ 
Slovene & \textbf{77.89} & 69.37 &  77.83 & \textbf{69.74}   \\ 
Spanish & 75.55 & 70.62 & \textbf{76.58}$^*$ & \textbf{71.60}$^*$ \\ 
Swedish &  75.00 & 65.66 & \textbf{75.62} & \textbf{65.95}  \\ 
\hline
Average & 82.67  & 76.49 & \textbf{83.27} & \textbf{76.78}  \\
\hline
\multicolumn{5}{c}{}\\
\end{tabular}
\centering
\vspace*{-2pt}\caption{Parsing accuracy (UAS and LAS, with punctuation) of the arc-swift  and NL-Covington parsers on CoNLL-XI (1st block) and CoNLL-X (2nd block) datasets. Best results for each language are  in bold. * indicates statistically significant improvements ($\alpha = .05$).}
\label{tab:results2}
\end{center}
\end{small}
\end{table}

Finally, we 
analyze
how our approach reduces the length of the transition sequence consumed by the original Covington parser. In Table~\ref{tab:results3} we report the transition sequence length per sentence used by the Covington and the NL-Covington algorithms to analyze each dataset from the same benchmark used for evaluating parsing accuracy. As 
seen in the table, NL-Covington produces notably shorter transition sequences than 
Covington,
with a reduction close to 50\% on average. 

\begin{table}
\begin{small}
\begin{center}
\centering
\begin{tabular}{@{\hskip 0pt}l|c|c@{\hskip 0pt}}
& Covington
& NL-Covington
\\
Language
& trans./sent. 
& trans./sent. 
\\
\hline
Arabic & 194.80 & \textbf{78.22} \\ 
Basque &  46.74 & \textbf{30.13}  \\ 
Catalan & 117.35 & \textbf{60.07} \\ 
Chinese & 19.12 & \textbf{14.95}   \\  
Czech & 60.62 & \textbf{33.03} \\ 
English & 78.01 &  \textbf{46.75} \\  
Greek &  89.23 & \textbf{48.77}  \\ 
Hungarian & 68.54 & \textbf{37.66}  \\ 
Italian & 63.67 & \textbf{40.93}   \\ 
Turkish & 53.53 & \textbf{30.08} \\ 
\hline
Bulgarian & 51.35 & \textbf{29.81} \\ 
Danish & 66.77 & \textbf{36.34} \\ 
Dutch & 42.78 & \textbf{28.93} \\ 
German &  61.16 & \textbf{31.89}  \\ 
Japanese & 24.30 & \textbf{16.11}   \\ 
Portuguese & 76.14 & \textbf{40.74} \\ 
Slovene & 56.15 & \textbf{31.79}    \\ 
Spanish & 109.70 & \textbf{55.28} \\ 
Swedish &  48.59 & \textbf{29.07}  \\ 
\hline
PTB-SD &  81.65 & \textbf{46.92}  \\ 
\hline
Average &  70.51 & \textbf{38.37}  \\
\hline
\multicolumn{3}{c}{}\\
\end{tabular}
\centering
\caption{
Average transitions executed per sentence (trans./sent.) when
analyzing each dataset by the original Covington and NL-Covington algorithms.}
\label{tab:results3}
\end{center}
\end{small}
\end{table}

\section{Conclusion}
We present a novel variant of the non-projective Covington transition-based parser by incorporating non-local transitions, reducing the length of transition sequences from $O(n^2)$ to $O(n)$. This system clearly outperforms the original Covington parser and achieves the highest accuracy on the WSJ Penn Treebank (Stanford Dependencies) obtained to date with greedy dependency parsing.

\section*{Acknowledgments}

This work has received funding from the European
Research Council (ERC), under the European
Union's Horizon 2020 research and innovation
programme (FASTPARSE, grant agreement No
714150), from the TELEPARES-UDC
(FFI2014-51978-C2-2-R) and ANSWER-ASAP (TIN2017-85160-C2-1-R) projects from MINECO, and from Xunta de Galicia (ED431B 2017/01).

\bibliography{main,twoplanaracl,bibliography}
\bibliographystyle{acl_natbib}

\appendix

\section{Model Details}
\label{app:modeldescription}

We provide more details of the neural network architecture used in this paper, which is taken from \newcite{Qi2017}.

The model consists of two blocks of 2-layered bidirectional long short-term memory (BiLSTM) networks \cite{Graves05} with 400 hidden units in each direction. The first block is used for POS tagging and the second one, for parsing. As the input of the tagging block, we use words represented as word embeddings, and BiLSTMs are employed to perform feature extraction. The resulting output is fed into a multi-layer perceptron (MLP), with a hidden layer of 100 rectified
linear  units  (ReLU), that provides a POS tag for each input token in a 32-dimensional representation. Word embeddings concatenated to these POS tag embeddings serve as input of the second block of BiLSTMs to undertake the parsing stage. Then, the output of the parsing block is fed into a MLP with two separate ReLU hidden layers (one for deriving the representation of the head, and the other for the dependency label) that, after being merged and by means of a softmax function, score all the feasible transitions, allowing to greedily choose and apply the highest-scoring one.

Moreover, we adapt the featurization process with biaffine combination described in \newcite{Qi2017} for the arc-swift system to be used on the original Covington and NL-Covington parsers. In particular, arc transitions are featurized by the concatenation of the representation of the head and dependent words of the arc to be created, the $\na$ transition is featurized by the rightmost word in $\lambda_1$ and the leftmost word in the buffer $B$ and, finally, for the $\sh$ transition only the leftmost word in $B$ is used. Unlike \newcite{Qi2017} do for baseline parsers, we do not use the featurization method detailed in \newcite{Kiperwasser2016}\footnote{For instance, \newcite{Kiperwasser2016} featurize all transitions of the arc-eager parser in the same way by concatenating the representations of the top 3 words on the stack and the leftmost word in the buffer.} for the original Covington parser, as we observed that this results in lower scores and then the comparison would be unfair in our case. We implement both systems under the same framework, with the original Covington parser represented as the NL-Covington system plus the $\na$ transition and with $k$ limited to 1. A thorough description of the model architecture and featurization mechanism can be found in \newcite{Qi2017}.

Our training setup is exactly the same used by \newcite{Qi2017}, training the models during 10 epochs for large datasets and 30 for small ones.
In addition, we initialize word embeddings with 100-dimensional GloVe vectors \cite{glove} for English and use 300-dimensional Facebook vectors \cite{facebookemb} for other languages. The other parameters of the neural network keep the same values. 

The parser's source code is freely available at \url{https://github.com/danifg/Non-Local-Covington}.

\end{document}